\documentclass[10pt,twocolumn,letterpaper]{article}

\usepackage{iccv}
\usepackage{times}
\usepackage{epsfig}
\usepackage{graphicx}
\usepackage{amsmath}
\usepackage{amssymb}

\usepackage[accsupp]{axessibility} % Improves PDF readability for those with disabilities.

\usepackage[breaklinks=true,colorlinks,bookmarks=false]{hyperref}

\iccvfinalcopy % *** Uncomment this line for the final submission

\def\iccvPaperID{11412} % *** Enter the ICCV Paper ID here

\ificcvfinal\fi

\begin{document}

\title{Sentence Attention Blocks for Answer Grounding}

\author{Seyedalireza Khoshsirat \qquad Chandra Kambhamettu\\
Video/Image Modeling and Synthesis (VIMS) Lab, University of Delaware\\
{\tt\small \{alireza, chandrak\}@udel.edu}
}

\maketitle
\ificcvfinal\thispagestyle{empty}\fi

\begin{abstract}
Answer grounding is the task of locating relevant visual evidence for the Visual Question Answering task.
While a wide variety of attention methods have been introduced for this task, they suffer from the following three problems: designs that do not allow the usage of pre-trained networks and do not benefit from large data pre-training, custom designs that are not based on well-grounded previous designs, therefore limiting the learning power of the network, or complicated designs that make it challenging to re-implement or improve them.
In this paper, we propose a novel architectural block, which we term Sentence Attention Block, to solve these problems.
The proposed block re-calibrates channel-wise image feature-maps by explicitly modeling inter-dependencies between the image feature-maps and sentence embedding.
We visually demonstrate how this block filters out irrelevant feature-maps channels based on sentence embedding.
We start our design with a well-known attention method, and by making minor modifications, we improve the results to achieve state-of-the-art accuracy.
The flexibility of our method makes it easy to use different pre-trained backbone networks, and its simplicity makes it easy to understand and be re-implemented.
We demonstrate the effectiveness of our method on the TextVQA-X, VQS, VQA-X, and VizWiz-VQA-Grounding datasets.
We perform multiple ablation studies to show the effectiveness of our design choices.
\end{abstract}

\section{Introduction}
\subsection{Visual Question Answering}
Visual Question Answering (VQA) systems try to accurately answer natural language questions regarding an input image \cite{antol2015vqa}.
This topic aims at developing systems that can communicate effectively about an image in natural language and comprehend the contents of images similar to humans.

\subsection{Answer Grounding}
The answer grounding task is defined as detecting the pixels that can provide evidence for the answer to a given question regarding an image \cite{chen2022grounding}.
In other words, the task is to return the image regions used to arrive at the answer for a given visual question (question-image pair) with an answer.
Although the VQA community has made significant progress, the best-performing systems are complicated black-box models, raising concerns about whether their answer reasoning is based on correct visual evidence.
By understanding the reasoning mechanism of the model, we can evaluate the quality of answers, improve model performance, and provide explanations for end-users.
To address this problem, answer grounding has been introduced into VQA systems, which requires the model to locate relevant image regions as well as answer visual questions.
\par
By providing answer groundings in response to visual questions, numerous applications become possible.
First, they allow for evaluating whether a VQA model is reasoning correctly based on visual evidence.
This is useful as an explanation as well as for assisting developers with model debugging.
Second, answer grounding makes it possible to separate important regions from irrelevant background regions.
Given that non-professional users can mistakenly have private information in the background of their pictures, answer grounding is a useful tool to obfuscate the background for privacy preserving.
Third, suppose a service can magnify the relevant visual evidence. In that case, users will be able to discover the needed information in less time.
This is useful in part because VQA answers can be insufficient sometimes.

\subsection{Multimodal Deep Learning}
By definition, the VQA and answer grounding tasks are multimodal tasks since a method for these tasks should be able to process and correlate two different modalities.
\par
\textbf{Multimodal Joint-Embedding Models}: These models merge and learn representations of multiple modalities in a joint feature space.
Joint-embeddings underpin the building of a lot of cross-modal methods as they can bridge the gap between different modalities.
In this joint space, the distance of different points is equivalent to the semantic distance between their corresponding original inputs.
\par
\textbf{Multimodal Attention-based Models}: The main objective of the attention mechanism is to design systems that use local features of image or text, extract features from different regions, and assign priorities to them.
The attention portion in an image determines salient regions.
Then the language generation component focuses more on those salient regions for additional processing \cite{summaira2021recent,bayoudh2022survey,hosseini2022application,khoshsirat2022semantic,maserat201743}.

\subsection{Attention Mechanism}
In deep learning, attention is a mechanism that mimics cognitive attention.
The goal is to enhance the essential features of the input data and vanish out the rest.
Attention methods can be classified into two classes based on their inputs: self-attention and cross-attention.
Self-attention (also known as intra-attention) is a type of attention that quantifies the interdependence within the elements of a single input.
At the same time, cross-attention (also known as inter-attention) finds the interdependence across two or more inputs \cite{vaswani2017attention,khoshsirat2023empowering,khoshsiratembedding}.
Usually, cross-attention methods are used for multimodal inputs \cite{wei2020multi}.
Cross-attention models first process individual modalities using modality-specific encoders, then the encoded features are fed into cross-attention modules.
\par
The Squeeze-and-Excitation method \cite{hu2018squeeze} is a channel-wise self-attention mechanism widely used in classification networks \cite{tan2019efficientnet, tan2021efficientnetv2}.
It consists of a global average pooling of the input, followed by two linear layers with an interleaved non-linearity and a sigmoid function.
Concretely, the output of this method is:
\begin{equation}
 \sigma (FC(RELU(FC(g\_avg\_pool(\mathbf{X}))))) \times \mathbf{X}
\end{equation}
where $\mathbf{X}$ is the feature-maps.
This module aims to dynamically focus on more important channels, essentially increasing the importance of specific channels over others.
This is accomplished by scaling the more important channels by a higher value.
While many feature descriptors exist to reduce the spatial dimensions of the feature maps to a singular value, this module uses average pooling to keep the required computation low.
The next part of the module maps the scaling weights using a Multi-Layer Perceptron (MLP) with a bottleneck structure.
The result values are scaled to a range of 0-1 by passing them through a sigmoid activation layer.
Afterward, using a common broadcasted element-wise multiplication, the output is applied directly to the input.
This multiplication scales each input feature-maps channel with its corresponding weight learned from the MLP in the Excitation module.
\par
The following is the summary of our contributions:
\begin{itemize}
\item We present a novel attention module based on the Squeeze-and-Excitation method for the answer grounding task.
\item We evaluate our method on common datasets and show that it achieves new state-of-the-art results.
\item We compare our design with the top-performing networks.
\item We perform multiple ablation studies to learn more about this network.
\end{itemize}

\section{Related Work}
\subsection{Attention Methods for Answer Grounding}
Models based on different attention methods have been actively explored for the answer grounding task.
\par
MAC-Caps \cite{urooj2021found} is based on MAC \cite{hudson2018compositional} which has a recurrent reasoning architecture that performs \textit{T} reasoning steps to answer the question.
At each reasoning step, MAC uses an attention block to read from image features and writing memory.
MAC-Caps adds capsule layers on top of the convolutional layers to obtain visual capsules from the feature-maps.
\par
Att-MCB \cite{riquelme2020explaining} is designed to use a Knowledge Base.
It is composed of nine modules, out of which two modules use an attention mechanism.
One module uses an attention mechanism to generate visual explanations, and the other one to consume and point out relevant information from the Knowledge Base.
\par
A multi-grained attention method is introduced in \cite{huang2019multi}.
It consists of two types of object-level groundings to explore fine-grained information, and a more sophisticated language model for better question representation.
A Word-Label Matching attention vector that indicates the weight that should be given to each of the \textit{K} objects in the image is computed in terms of the semantic similarity between the category labels of the objects and the words in the question.
A word-object matching module is exploited to evaluate how likely a question word matches a visual object. A Sentence-Object attention module is used to capture the global semantics of the whole sentence to guide the focus on relevant objects.
\par
Multiple custom-designed attention modules are used in \cite{zhang2019interpretable} to generate an attention map for the given image and question pair.
Questions are tokenized and passed through an embedding layer followed by an LSTM layer to generate the question features.
An Image Attention Supervision Module is used as an auxiliary classification task; that is, the ground-truth visual grounding labels are used to guide the model to focus on significant parts of the image to answer each question.
\par
A so-called "accumulated attention" mechanism is introduced in \cite{deng2018visual}.
It consists of three attention modules, one for the input text, one for the objects, and one for the whole image.
All three modules have a similar structure of a linear layer followed by a Softmax function.

\begin{figure}[t]
\begin{center}
  \includegraphics[width=0.9\linewidth]{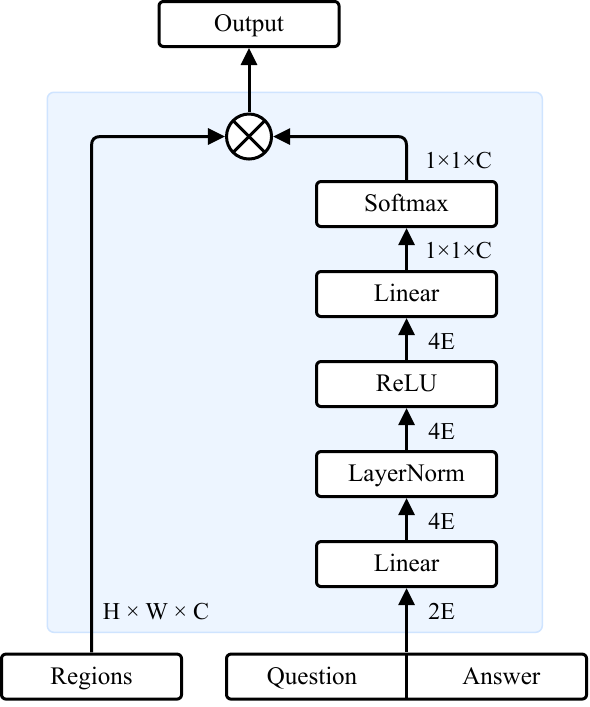}
\end{center}
\caption{Our proposed sentence attention block.
$C$ denotes the number of channels in image feature-maps, and $E$ represents the embedding size of the sentence encoder network.
Question and answer embeddings are concatenated into one vector.}
\label{fig:block}
\end{figure}

\begin{figure*}[t]
\begin{center}
  \includegraphics[width=0.9\linewidth]{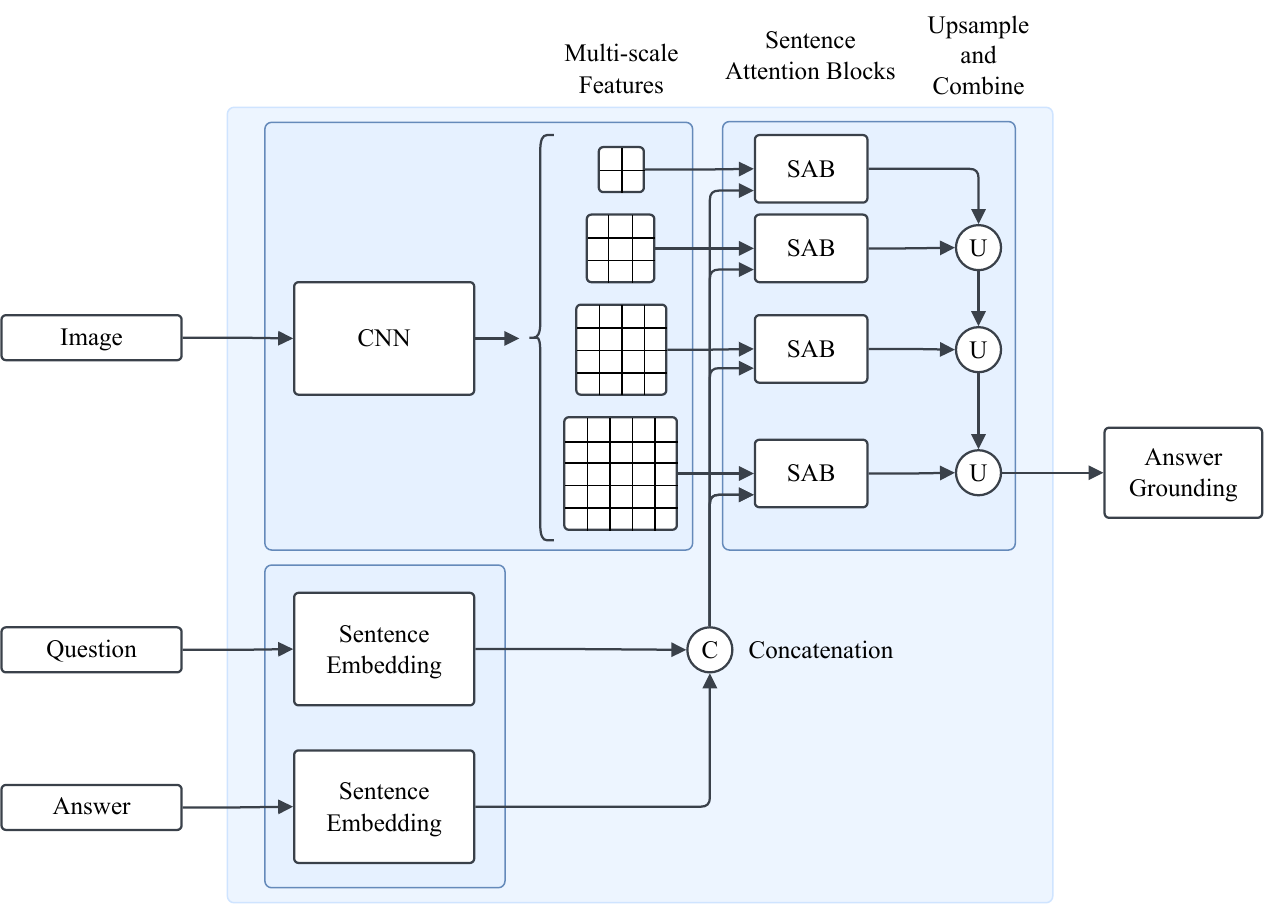}
\end{center}
\caption{The complete proposed network.
The multi-scale channel-wise image features contain different regions that are filtered by our Sentence Attention Block (SAB) based on the question and answer embeddings.}
\label{fig:network}
\end{figure*}

\subsection{Multimodal Attention Methods}
Since its introduction, the attention mechanism has been widely adopted in the computer vision community due to its special capabilities for many multimodal applications.
\par
In \cite{hori2017attention}, a multimodal attention model is proposed based on the encoder-decoder networks and RNNs for video captioning and sentence generation.
In particular, the multimodal attention framework combined image, audio, and motion features by selecting each modality's most relevant context vector.
\par
In \cite{yang2016stacked}, utilizing stacked attention networks is suggested to look for image regions that correlate with a query answer and pinpoint representative features of a given question more accurately.
\par
More recently, in \cite{guo2020normalized}, a normalized variant of the self-attention mechanism, named normalized self-attention (NSA), is introduced.
NSA seeks to encode and decode the image and caption features and normalize the distribution of internal activations during training.
\par
For the video question answering task, attention and memory mechanisms are used in \cite{fan2019heterogeneous} to efficiently learn visual features and the semantic correlations that precisely answer questions.

\subsection{Answer Grounding Methods}
The answer grounding task has also been studied under different names, such as grounded VQA and visual explanation for VQA.
\par
The VinVL \cite{zhang2021vinvl} method is based on OSCAR \cite{li2020oscar} which proposes a novel vision-language pre-training that incorporates image anchor points as input, taken from an object detection network.
VinVL improves the visual representations by using a bigger visual model pre-trained on larger training corpora that combine multiple annotated object detection datasets.
\par
LXMERT \cite{tan2019lxmert} presents a cross-modality framework for learning the connections between vision and language.
The authors build a large-scale Transformer model that consists of three encoders: an object relationship encoder, a language encoder, and a cross-modality encoder.
This model is then pre-trained on a large-scale dataset of image-and-sentence pairs.
\par
MTXNet \cite{rao2021first} is an end-to-end trainable architecture that generates multimodal explanations focusing on the text in the image.
It contains a Graph Attention Network, and each object location and OCR token is treated as a node in the graph.
It also contains a multimodal transformer that operates on three modalities: question words, visual objects, and OCR tokens.
\par
U-CAM \cite{patro2019u} is a method that uses gradient-based uncertainty estimates to provide visual attention maps.
The certainty measure is improved by adding the certainty gradients to the existing standard Cross-Entropy loss gradients for training the model during back-propagation.

\section{Method}
Current attention methods have one or more of the following three problems:
\begin{itemize}
\item The attention mechanism is not based on well-known and well-tested methods \cite{riquelme2020explaining,tan2019lxmert,patro2019u}.
Well-established methods have been evaluated on many different tasks and datasets, especially, they perform well on large-scale datasets.
Their design details have been actively studied, and their advantages and disadvantages are visible to the users.
\item The complicated design makes it challenging to re-implement them on other platforms \cite{riquelme2020explaining, urooj2021found, rao2021first}.
Recently, many applications have been made based on deep neural networks that are expected to be used by the end-users of different platforms.
For example, answer grounding methods are helpful in assistive technologies for people with vision impairments \cite{chen2022grounding}.
In order to use such technologies offline, it is crucial to implement the method for each platform.
\item The custom design does not allow the usage of pre-trained models \cite{urooj2021found,zhang2021vinvl,patro2020robust,park2018multimodal}.
Answer grounding datasets are less diverse and have fewer samples than the classification datasets.
Using pre-trained models can improve the accuracy and make the model more robust to noise and unseen samples.
\end{itemize}
This paper proposes an attention block that avoids the mentioned problems.
We design an attention block based on well-established and well-studied methods.
It has a simple design to make it easy to re-implement and uses pre-trained models to achieve maximum accuracy.
\par
Our proposed method has three main parts: region proposal, sentence embedding, and attention fusion.
\subsection{Region Proposal}
This module's primary goal is to process the image and generate all the possible region candidates.
Unlike Region Proposal Networks \cite{ren2015faster} which generate anchor-based bounding boxes, this module generates dense regions (segmentations).
\par
A pre-trained classification network is used as the backbone.
However, the final classifier is removed to get the multi-scale feature-maps.
During the end-to-end training of the proposed method, the backbone network learns all the candidate regions.
Later, these candidate regions are filtered and combined by the proposed Sentence Attention Blocks.
In Section \ref{sec:region-proposals}, we show that the backbone network works as expected by visualizing the feature-maps.

\subsection{Sentence Embedding}
Since the input questions and answers have variable lengths, they are processed by a sentence embedding network.
A pre-trained sentence encoder is employed to encode the given question and answer to a vector space separately.
The result is two vectors, one for the question and one for the answer.
The two vectors are concatenated and passed to the Sentence Attention Blocks as one vector.

\begin{table}[t]
\begin{center}
\setlength\tabcolsep{20pt}
\begin{tabular}{l|l}
Method & Mean IoU \\
\hline
MTXNet \cite{rao2021first} & 18.9\\
LXMERT \cite{tan2019lxmert} & 24.9\\
VinVL \cite{zhang2021vinvl} & 25.6\\
MAC-Caps \cite{urooj2021found} & 26.1\\
\hline
\textbf{Ours} & \textbf{29.0 \textcolor{green}{(+2.9)}}\\
\hline
\end{tabular}
\end{center}
\caption{Comparison results on the test set of the TextVQA-X dataset \cite{rao2021first}.}
\label{tab:textvqa-x-results}
\end{table}

\subsection{Attention Fusion} \label{sec:attention-fusion}
Given multiple potential regions for an image, the answer grounding task reduces to choosing and/or combining these regions based on the given question and/or answer.
The multi-scale channel-wise image features from the backbone network contain candidate regions.
The goal is to filter and combine these regions based on the question and answer vector from the sentence embedding network.
Towards this goal, we design an attention block based on the Squeeze-and-Excitation (SE) block \cite{hu2018squeeze}.
\par
We start with an SE block and make four modifications as follows:
\begin{enumerate}
\item The first modification is changing it from self-attention to cross-attention.
In other words, unlike a self-attention block, the proposed block pays attention to an external source: the encoded question and answer.
\item The second modification is adding a normalization layer.
Empirically we found that adding a normalization layer improves the accuracy.
Similar to \cite{vaswani2017attention}, we use LayerNorm since it performs better on sequences.
\item The third modification is replacing the sigmoid function with softmax.
Since the goal is to force the model to choose a few of the channel-wise image features, the sigmoid function is replaced with softmax.
\item The last modification is inverting the bottleneck.
Instead of shrinking the embedding space, we expand the question and answer embeddings by a factor of two.
\end{enumerate}
In Section \ref{sec:design-modifications}, the effect of each modification on the network accuracy is shown.
Figure \ref{fig:block} depicts the final structure of the proposed block.
Precisely, the proposed attention block is as follows:
\begin{equation}
 Softmax(FC(ReLU(LN(FC(\mathbf{QA}))))) \times \mathbf{R}_i
\end{equation}
where $FC$ is a single-layer perceptron, $LN$ is LayerNorm, $\mathbf{QA}$ is the joint question and answer embeddings, and $\mathbf{R}_i$ is the image regions at scale $i$.
For each scale, a separate instance of this attention block is created so that each block learns special weights for its corresponding feature-map scale.
\par
After computing the attention block for all the scales, all the outputs at different scales are combined into one output.
Starting from the lowest resolution output, a $1 \times 1$ convolution is used to convert the number of channels to the number of channels of the next lowest resolution output.
Then, Bilinear interpolation is used for upsampling the output feature-map by a factor of 2.
Finally, the output is added to the next lowest resolution output.
This process is repeated until no other output is left.
In the end, a $1 \times 1$ convolution is used to reduce the number of channels of the final output to two.
Therefore, the final output has two channels representing answer grounding and its background.
We use Argmax to output a binary image.
\par
The complete design of the proposed method is depicted in Figure \ref{fig:network}.

\begin{table}[t]
\begin{center}
\setlength\tabcolsep{20pt}
\begin{tabular}{l|l}
Method & Mean IoU \\
\hline
DeconvNet \cite{gan2017vqs} & 29.8\\
Mask Aggregation \cite{gan2017vqs} & 32.6\\
LXMERT \cite{tan2019lxmert} & 33.3\\
VinVL \cite{zhang2021vinvl} & 33.9\\
MAC-Caps \cite{urooj2021found} & 34.3\\
\hline
\textbf{Ours} & \textbf{36.8 \textcolor{green}{(+2.5)}}\\
\hline
\end{tabular}
\end{center}
\caption{Comparison results on the test set of the VQS dataset \cite{gan2017vqs}.}
\label{tab:vqs-results}
\end{table}

\subsection{Comparison to Existing Methods}
We compare the design of our proposed method to the top-performing existing methods for the answer grounding task.
\par
MAC-Caps \cite{urooj2021found} is one of the best-performing networks that currently exist.
This network is based on MAC \cite{hudson2018compositional} which employs attention blocks in a recurrent reasoning architecture.
MAC-Caps adds capsule layers to process the feature-maps further.
Although similar to our method, MAC-Caps employs an attention mechanism.
However, the recurrent design brings the usual problems of RNNs, such as gradient vanishing/exploding, longer training time, and not performing well on long sequences.
Since our proposed method does not have a recurrent design, it avoids such problems.
\par
VinVL \cite{zhang2021vinvl} is another top-performing network for the answer grounding task.
This method is based on OSCAR \cite{li2020oscar} which incorporates image anchor points as inputs.
VinVL improves the visual representations by using a bigger visual model pre-trained on larger training corpora that combine multiple annotated object detection datasets.
This method relies on object detection networks to provide the anchor points, which adds an extra task for the network to learn.
Also, using large generic corpora may not improve the accuracy for special datasets such as VizWiz-VQA-Grounding \cite{chen2022grounding}.
In contrast, our proposed method relies only on the feature-maps and does not define another task.
\par
Att-MCB \cite{riquelme2020explaining} is the state-of-the-art network on the VQA-X dataset \cite{park2018multimodal}.
This method is designed to use a Knowledge Base.
It comprises nine modules and four loss functions, making it a tedious design.
This complicated design is resource-demanding and makes the training sensitive to hyper-parameters.
Our proposed method comprises three main modules and one loss function.
This minimalistic design helps the gradient flow during back-propagation, and the training is less sensitive to hyper-parameters.

\begin{table}[t]
\begin{center}
\setlength\tabcolsep{15pt}
\begin{tabular}{l|l}
Method & Rank Correlation \\
\hline
PJ-X \cite{park2018multimodal} & 0.342\\
CCM \cite{patro2020robust} & 0.368\\
U-CAM \cite{patro2019u} & 0.372\\
VinVL \cite{zhang2021vinvl} & 0.373\\
Att-MFH \cite{riquelme2020explaining} & 0.376\\
MAC-Caps \cite{urooj2021found} & 0.389\\
Att-MCB \cite{riquelme2020explaining} & 0.396\\
\hline
\textbf{Ours} & \textbf{0.421 \textcolor{green}{(+0.025)}}\\
\hline
\end{tabular}
\end{center}
\caption{Comparison results on the test set of the VQA-X dataset \cite{park2018multimodal}.}
\label{tab:vqa-x-results}
\end{table}

\section{Experiments}
We train and test our proposed method on four answer grounding datasets.
We use the same setup for all of the experiments.

\subsection{Setup} \label{sec:setup}
We use a pre-trained EfficientNet \cite{tan2019efficientnet} to extract the multi-scale image features and we use a MiniLMv2 \cite{wang2020minilmv2} sentence encoder network to encode the questions.
In Section \ref{sec:backbone-networks}, we perform an ablation study on different backbones.
we use BEiT \cite{wang2022image} to generate answers for the datasets where the answers are not publicly available.
We use RMI \cite{zhao2019region} with its default settings along with cross-entropy.
Therefore, the overall loss function to minimize is:
\begin{equation}
\begin{split}
 \mathcal{L}_{all}(y, p) & = \frac{1}{B} \sum_{b=1}^{B} \lambda \mathcal{L}_{ce}(y^{(b)},p^{(b)}) \\ & + (1-\lambda) \mathcal{L}_{rmi}(y^{(b)},p^{(b)})
\end{split}
\end{equation}
where $\lambda \in [0, 1]$ is a weight factor, $B$ denotes the number of samples in a mini-batch, $\mathcal{L}_{ce}(y^{(b)},p^{(b)})$ is the standard cross entropy loss between the $b$-th sample and its corresponding prediction, respectively, and $\mathcal{L}_{rmi}(y^{(b)},p^{(b)})$ is the RMI loss as follows:
\begin{equation}
 \mathcal{L}_{rmi}(\mathbf{Y}, \mathbf{P}) = \sum_{c=1}^{C} \frac{1}{2d} Trace(log(\mathbf{M}))
\end{equation}
and,
\begin{equation}
 \mathbf{M} = \mathbf{\Sigma}_\mathbf{Y} - Cov(\mathbf{Y},\mathbf{P})(\mathbf{\Sigma}_\mathbf{P}^{-1})^T Cov(\mathbf{Y},\mathbf{P})^T
\end{equation}
where $\mathbf{Y}=y^{(b)}$, $\mathbf{P}=p^{(b)}$, $\mathbf{M} \in \mathbb{R}^{d \times d}$, $d$ denotes the number of pixels, $C$ is the number of object classes, here $C=2$ for the answer grounding and its background, $\mathbf{\Sigma}_\mathbf{Y}$ is the variance matrix of Y, and $Cov(\mathbf{Y},\mathbf{P})$ is the covariance matrix of $\mathbf{Y}$ and $\mathbf{P}$.
Similar to \cite{zhao2019region}, we set $\lambda=0.5$.
\par
We use AdamW optimizer \cite{loshchilov2017decoupled} with a weight decay of 0.05 and batch size of 16.
We apply the "polynomial" learning rate policy with a poly exponent of 0.9 and an initial learning rate of 0.0001.
Synchronized batch normalization is used across multiple GPUs.
We use RandAugment \cite{cubuk2020randaugment} for data augmentation.

\begin{table}[t]
\begin{center}
\setlength\tabcolsep{15pt}
\begin{tabular}{l|l}
Team & Mean IoU \\
\hline
MindX & 66.9\\
MGTV & 69.7\\
hsslab\_inspur & 70.1\\
GroundTruth & 70.3\\
Aurora & 70.6\\
\hline
\textbf{SAB (Ours)} & \textbf{72.4 \textcolor{green}{(+1.8)}}\\
\hline
\end{tabular}
\end{center}
\caption{The top-performing teams on the VizWiz-VQA-Grounding Challenge 2022 leaderboard \cite{chen2022grounding}.
At the time of this writing, none of the methods have been published.}
\label{tab:vizwiz-leaderboard}
\end{table}

\begin{table}[t]
\begin{center}
\setlength\tabcolsep{10pt}
\begin{tabular}{l|cc}
Method & \# of params & inference time \\
\hline
LXMERT \cite{tan2019lxmert} & 153M & 56ms \\
VinVL \cite{zhang2021vinvl} & 406M & 69ms \\
MAC-Caps \cite{urooj2021found} & 80M & 54ms \\
\hline
Ours & 88M & \textbf{49ms} \\
\hline
\end{tabular}
\end{center}
\caption{Comparison of our proposed method to existing methods regarding the number of parameters and inference time.}
\label{tab:comparison}
\end{table}

\subsection{TextVQA-X} \label{sec:textvqa-x}
The TextVQA-X dataset \cite{rao2021first} is a subset of the TextVQA dataset \cite{singh2019towards} such that the answer groundings are generated through a manual segmentation annotation process.
TextVQA-X consists of 10,379 training images and 3,354 test images.
Since only a few models have been evaluated on this dataset, we train three more models for a fair comparison.
Similar to \cite{chen2022grounding}, we train top-performing methods that their code is publicly available; specifically, LXMERT \cite{tan2019lxmert}, VinVL \cite{zhang2021vinvl}, and MAC-Caps \cite{urooj2021found}.
We use the same setup as in \cite{chen2022grounding} to train the three models with the difference that we train the models on the TextVQA-X training set.
Similar to the existing methods, we use the test set to evaluate our method and report the mean IoU measure.

\subsection{VQS}
The VQS dataset \cite{gan2017vqs} builds upon the images, instance segmentation masks, and bounding boxes in COCO \cite{lin2014microsoft} and the questions and answers in the VQA dataset \cite{antol2015vqa}.
VQS contains 26,995 training, 5,000 validation, and 5,873 test images.
Similar to Section \ref{sec:textvqa-x}, we train top-performing methods that their code is publicly available; specifically, LXMERT \cite{tan2019lxmert}, VinVL \cite{zhang2021vinvl}, and MAC-Caps \cite{urooj2021found}.
We use the same setup as in \cite{chen2022grounding} to train the three models with the difference that we train the models on the VQS training set.
Similar to the existing methods, we use the test set to evaluate our method and report the mean IoU measure.

\subsection{VQA-X}
The VQA-X dataset \cite{park2018multimodal} does not have visual annotations for its training set.
It has visual annotations only for its validation and test sets.
Therefore, we train our method on the training set of the VQS dataset \cite{gan2017vqs}, which has related images to the VQA-X validation and test sets.
The images of both datasets are from the COCO dataset \cite{lin2014microsoft}.
The other methods are trained on different datasets, including Visual Genome \cite{krishna2017visual} and VQA v2.0 \cite{goyal2017making}.
The VQA-X dataset consists of 24,876 training, 1,431 validation, and 1,921 test images.
Similar to the other methods, we use the test set of the VQA-X dataset to evaluate our method and report the Rank Correlation as in \cite{park2018multimodal}.

\subsection{VizWiz-VQA-Grounding}
The VizWiz-VQA-Grounding dataset \cite{chen2022grounding} is based on the VizWiz-VQA dataset \cite{gurari2018vizwiz}, and the images and questions come from visually impaired people who shared them to request visual assistance in their day-to-day lives.
VizWiz-VQA-Grounding contains a total of 9,998 VQAs divided into 6,494/1,131/2,373 VQAs for training, validation, and testing.
The 2022 VizWiz-VQA-Grounding Challenge is designed around the aforementioned VizWiz-VQA-Grounding dataset.
This challenge was held recently, and at the time of this writing, current methods in this challenge have not been published yet.
Therefore, we evaluate our method using the VizWiz-VQA-Grounding Challenge online leaderboard \cite{chen2022grounding}.
The leaderboard evaluates all the methods against the VizWiz-VQA-Grounding test set.
The ground truth labels for the test set are not published yet.

\begin{table}[t]
\begin{center}
\setlength\tabcolsep{10pt}
\begin{tabular}{l|c}
Backbone & Mean IoU \\
\hline
ResNet-50x3 \cite{kolesnikov2020big} & 71.9\\
Swin Transformer V2-B \cite{liu2021swin} & 72.8\\
EfficientNetV2-L \cite{tan2021efficientnetv2} & 73.1\\
EfficientNet-B7 \cite{tan2019efficientnet} & \textbf{73.5}\\
\hline
RoBERTa \cite{liu2019roberta} & 72.8\\
MPNet \cite{song2020mpnet} & 73.1\\
MiniLMv2 \cite{wang2020minilmv2} & \textbf{73.5}\\
\hline
\end{tabular}
\end{center}
\caption{The ablation study of the pre-trained backbone networks.
\textbf{Top Section:} Using a MiniLMv2 as a fixed sentence embedding network and replacing the image feature extraction networks.
\textbf{Bottom Section:} Using a fixed image feature extraction network (EfficientNet-B7) and replacing the sentence embedding networks.}
\label{tab:backbone-networks}
\end{table}

\subsection{Results}
We evaluate our proposed method on four answer grounding datasets.
Tables \ref{tab:textvqa-x-results}, \ref{tab:vqs-results}, and \ref{tab:vqa-x-results} compare the results of our proposed method with the existing methods on the TextVQA-X, VQS, and VQA-X datasets respectively.
Our method achieves new state-of-the-art accuracy on three datasets.
More specifically, improvements of 2.9, 2.5, and 0.025 on the TextVQA-X, VQS, and VQA-X datasets, respectively.
\par
Table \ref{tab:vizwiz-leaderboard} lists the top-performing teams on the VizWiz-VQA-Grounding Challenge 2022 leaderboard \cite{chen2022grounding}.
At the time of this writing, our method holds first place with considerably higher accuracy than the other methods.
\par
Furthermore, Table \ref{tab:comparison} compares our proposed method to existing methods regarding the number of parameters and inference time.
The efficient design of our method makes it the fastest method.

\section{Ablation Studies}
In this section, we perform four ablation studies.
All of the experiments are done on the VizWiz-VQA-Grounding \cite{chen2022grounding} validation set and using the setup from Section \ref{sec:setup} unless otherwise stated.

\subsection{Design Modifications} \label{sec:design-modifications}
We started the design of our sentence attention block from the standard SE block and made four modifications as explained in Section \ref{sec:attention-fusion}.
In this section, we perform an ablation study on these modifications.
Table \ref{tab:modifications} shows the impact of each modification on the final accuracy.
These modifications are done cumulatively.
The first modification is changing from self-attention to cross-attention, which makes our baseline for this ablation study.
The second modification is adding a normalization layer.
This modification improves our baseline by 1.1 percent.
The third modification is replacing the Sigmoid function with Softmax.
This modification has the highest impact on our method, 1.5 percent.
And the fourth modification is the expansion of the embeddings.
This last modification brings an addition of 0.8 percent.

\begin{table}[t]
\begin{center}
\setlength\tabcolsep{10pt}
\begin{tabular}{l|l}
Modification & Mean IoU \\
\hline
Switching to cross-attention & 70.1\\
Adding normalization layer & 71.2 \textcolor{green}{(+1.1)}\\
Switching to Softmax & 72.7 \textcolor{green}{(+1.5)}\\
Expansion of the embeddings & \textbf{73.5 \textcolor{green}{(+0.8)}}\\
\hline
\end{tabular}
\end{center}
\caption{The results of our ablation study of the design modifications.
The modifications are done cumulatively from top to bottom.}
\label{tab:modifications}
\end{table}

\subsection{Backbone Networks} \label{sec:backbone-networks}
The flexibility of our method enables us to use a variety of pre-trained networks.
In this ablation study, we compare and analyze different pre-trained networks.
To compare the image backbone networks, we keep the sentence embedding network fixed and change the image backbone network.
Similarly, to compare the sentence embedding networks, we keep the image backbone network with the highest accuracy and change the sentence embedding network.
We use well-known and well-performing networks whose pre-trained weights are publicly available.
Specifically, we use EfficientNet \cite{tan2019efficientnet}, EfficientNetV2 \cite{tan2021efficientnetv2}, Swin Transformer V2 \cite{liu2021swin}, and ResNet-50x3 \cite{kolesnikov2020big} for image feature extraction and MiniLMv2 \cite{wang2020minilmv2}, MPNet \cite{song2020mpnet}, and RoBERTa \cite{liu2019roberta} for sentence embedding.
Table \ref{tab:backbone-networks} shows the results of this ablation study.
The highest accuracy is achieved by using EfficientNet-B7 and MiniLMv2.
Since the EfficientNetV2-L reduces the input resolution more aggressively than EfficientNet-B7, it performs worse.
Transformer networks usually perform well on large datasets.
However, since VizWiz-VQA-Grounding is not a large dataset, Swin Transformer V2 does not achieve the highest accuracy.
The ResNet-50x3 is pre-trained on ImageNet-21k, but it achieves the least accuracy in this experiment.

\subsection{Region Proposals} \label{sec:region-proposals}
Feature-map visualization will provide insight into the internal representations of each layer in a network for a specific input image.
\par
\begin{figure}[t]
\begin{center}
  \includegraphics[width=0.9\linewidth]{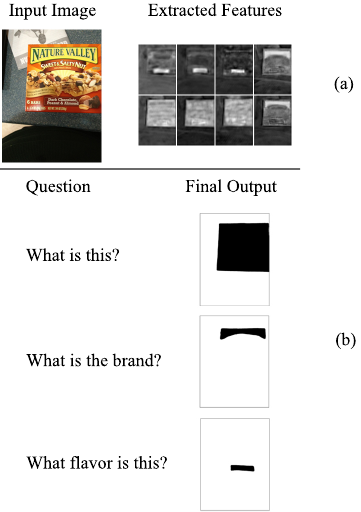}
\end{center}
\caption{\textbf{(a)} Eight manually-selected sample feature-maps for an image.
These feature-maps are the output of the feature extraction backbone network, before feeding them to the sentence attention block.
The network has learned to output potential regions in different channels.
\textbf{(b)} The final output of our complete network for the same image in (a), given different questions and answers (for brevity, the answers are not shown).
This figure shows that our proposed attention block can filter the channel-wise feature-maps based on the input questions and answers.}
\label{fig:regions}
\end{figure}

This ablation study aims to show that the image backbone network has learned the potential regions for an image, which are then processed by the proposed sentence attention block.
To this aim, we visualize the final output feature-maps of the image backbone network, before feeding them to an attention block.
We use the image backbone of our fully trained network on the VizWiz dataset.
Figure \ref{fig:regions} (a) shows a few manually-chosen feature-maps for a sample image.
These feature-maps show that after training, the backbone network has been able to learn the potential important regions of images.
Having all the candidate regions, the task of the proposed attention block is to filter the irrelevant regions to the question and answer.
Figure \ref{fig:regions} (b) shows the final output of the complete network for the same image in (a), using different questions and answers (for brevity, the answers are not shown).
This figure shows that our proposed block is able to filter the channel-wise feature-maps based on the input questions and answers.

\begin{table}
\begin{center}
\setlength\tabcolsep{15pt}
\begin{tabular}{l|l}
Scale & Mean IoU \\
\hline
1/32 & 71.4\\
1/32 + 1/16 & 72.5 \textcolor{green}{(+1.1)}\\
1/32 + 1/16 + 1/8 & 73.1 \textcolor{green}{(+0.6)}\\
1/32 + 1/16 + 1/8 + 1/4 & \textbf{73.5 \textcolor{green}{(+0.4)}}\\
\hline
\end{tabular}
\end{center}
\caption{The gradual effect of multi-scale fusion.
From top to bottom, the first row uses only the lowest resolution feature-map, and each row adds higher resolution feature-maps.
Each scale is the ratio of the feature-map size to the input image.}
\label{tab:multi-scale-fusions}
\end{table}

\subsection{Multi-scale Fusions}
There is a trade-off in the answer grounding task such that some images are best handled at lower inference resolution and others better handled at higher inference resolution.
Fine details, such as the edges of objects or thin structures, are often better predicted with scaled-up images.
At the same time, the prediction of large structures, which requires more global context, is often done better at scaled-down images because the network’s receptive field can observe more of the necessary context.
\par
This ablation study shows the impact of adding the extra scales on the accuracy.
Table \ref{tab:multi-scale-fusions} shows the results of this study.
From top row to bottom, we start with only one scale and one attention block and gradually add higher resolution feature-maps.
The lowest resolution is the final output feature-map of the backbone network.

\section{Conclusions}
In the study, a new architectural block termed the Sentence Attention Block was introduced to address certain challenges.
This block recalibrates channel-wise image feature-maps by modeling inter-dependencies between image feature-maps and sentence embedding.
The design began with a recognized attention method, and with minor adjustments, enhanced results were achieved, reaching top-tier accuracy.
We showed the block's ability to filter out irrelevant feature-map channels based on sentence embedding.
The method's adaptability allows for various pre-trained backbone networks, and its straightforward nature facilitates comprehension and re-implementation.
The method's efficacy was demonstrated on the TextVQA-X, VQS, VQA-X, and VizWiz-VQA-Grounding datasets, and several ablation studies highlighted the validity of the design decisions.

{\small
\bibliographystyle{ieee_fullname}
\bibliography{references}
}

\end{document}